\pdfoutput=1
\documentclass[11pt]{article}

\usepackage[preprint]{acl}
\usepackage[greek,english]{babel}
\usepackage{url}
\usepackage{times}
\usepackage{latexsym}
\usepackage{placeins}

\usepackage[T5]{fontenc}
\usepackage[utf8]{inputenc}

\usepackage{microtype}

\usepackage{inconsolata}
\usepackage[T1]{fontenc}

\usepackage{csquotes}
\usepackage{amsmath}
\usepackage[capitalise]{cleveref}
\usepackage{array}
\usepackage{graphicx}
\usepackage{multirow}
\usepackage{booktabs}
\usepackage{arydshln} 
\usepackage{threeparttable}
\usepackage[format=hang]{caption}
\usepackage{siunitx}
\usepackage{float}
\usepackage{adjustbox}
\title{
Heidelberg-Boston @ SIGTYP 2024 Shared Task:\\
Enhancing Low-Resource Language Analysis With Character-Aware Hierarchical Transformers
}

\author{Frederick Riemenschneider\thanks{Equal contribution.} \\
  Dept. of Computational Linguistics\\
  Heidelberg University, Germany\\
  \texttt{riemenschneider@cl.uni-heidelberg.de} \\\And
  Kevin Krahn\footnotemark[1] \\
  Dept. of Computer Science \\
  Sattler College, USA \\
  \texttt{kevin.krahn24@sattler.edu} \\}

\DeclareTextSymbolDefault{\ohorn}{T5}
\DeclareTextSymbolDefault{\uhorn}{T5}

\begin{document}
\maketitle
\begin{abstract}
Historical languages present unique challenges to the NLP community, with one prominent hurdle being the limited resources available in their closed corpora. This work describes our submission to the constrained subtask of the SIGTYP 2024 shared task, focusing on PoS tagging, morphological tagging, and lemmatization for 13 historical languages. For PoS and morphological tagging we adapt a hierarchical tokenization method from \citet{sun-etal-2023-characters} and combine it with the advantages of the DeBERTa-V3 architecture, enabling our models to efficiently learn from every character in the training data. We also demonstrate the effectiveness of character-level T5 models on the lemmatization task. Pre-trained from scratch with limited data, our models achieved first place in the constrained subtask, nearly reaching the performance levels of the unconstrained task's winner.
Our code is available at \url{https://github.com/bowphs/SIGTYP-2024-hierarchical-transformers}.
\end{abstract}

\section{Introduction}
Unlike modern languages, historical languages come with a notable challenge: their corpora are closed, meaning they cannot grow any further. This situation often puts researchers of historical languages in a low-resource setting, requiring tailored strategies to handle language processing and analysis effectively \citep{johnson-etal-2021-classical}.

In this paper, we focus on identifying the most efficient methods for extracting information from small corpora. In such a scenario, the main hurdle is not computational capacity, but learning to extract the maximal amount of information from our existing data.

To evaluate this, the SIGTYP 2024 shared task offers a targeted platform centering on the evaluation of embeddings and systems for historical languages. 
This task provides a systematic testbed for researchers, allowing us to assess our methodologies in a controlled evaluation setting for historical language processing.

For the constrained subtask, participants received annotated datasets for 13 historical languages sourced from Universal Dependencies \citep{11234/1-5150}, along with data for Old Hungarian that adheres to similar annotation standards \citep{simon2014, mgtsz2018}. These languages represent four distinct language families and employ six different scripts, which ensures a high level of diversity. The rules imposed in this subtask strictly forbid the use of pre-trained models and limit training exclusively to the data of the specified language. This restriction not only ensures full comparability of the applied methods, it also inhibits any cross-lingual transfer effects.

We demonstrate that, even in these resource-limited settings, it is feasible to achieve high performance using monolingual models. Our models are exclusively pre-trained on very small corpora, leveraging recent advances in pre-training language models. Our submission was recognized as the winner in the constrained task. Notably, it also delivered competitive results in comparison to the submissions in the unconstrained task, where the use of additional data was permitted. This highlights the strength of our approach, even within a more restricted data environment.

\section{Pre-trained Language Models for Ancient and Historical Languages}

\begin{table*}[t]
    \centering
    \begin{tabular}{lrrrrrrrrrrrrr}
        \toprule
        \textbf{Language:} & chu & cop & fro & got & grc & hbo & isl & lat & latm & lzh & ohu & orv & san\\
        \midrule
        \textbf{Vocab Size:} & 196 & 82 & 106 & 87 & 242 & 94 & 150 & 188 & 111 & 5714 & 166 & 222 & 62\\
        \bottomrule
    \end{tabular}
    \caption{Character vocabulary sizes (including special tokens). See \cref{sec:data} for language identifiers. }
    \label{tab:vocab}
\end{table*}

Much of the previous work on Pre-trained Language Models (PLMs) for ancient and historical languages has focused on cross-lingual transfer learning techniques \citep{krahn-etal-2023-sentence, singh-etal-2021-pilot,yamshchikov-etal-2022-bert,yousef-etal-2022-automatic} or languages with relatively large corpora compared to most historical languages, such as Ancient Greek and Latin \citep{riemenschneider-frank-2023-exploring, bamman2020latin}. In this work, we are interested in maximizing performance in more resource-limited environments while training exclusively on monolingual data.

\subsection{Representing Words and Characters}

Low-resource historical languages present several challenges for subword tokenizers which are typically used by PLMs. Given that our downstream tasks require predictions at the world level, it is important that the model learns good word representations in training.
At the same time, it is important to obtain good character representations because characters carry important morphological information. In small-scale training corpora, subword tokenizers are ineffective at capturing information at both the word and character levels, as shown in prior work \citep{clark-etal-2022-canine,kann-etal-2018-character}. As a result, it is difficult for a model to learn meaningful representations for rare tokens, which can be completely opaque to the model with respect to the characters they contain.

Adopting a character-based tokenizer would solve many of these problems, but as a downside would result in a much higher number of input tokens. Critically, the computational requirements of self-attention grow quadratically with sequence length, making training and inference time prohibitive or requiring truncated input sequences.

For these reasons, we adopt a solution for our encoder-only models that combines the advantages of word- and character-level representations. We base our architecture on the Hierarchical Pre-trained Language Model (HLM)
architecture recently proposed by \citet{sun-etal-2023-characters}, which solves many of our problems. HLM is a hierarchical two-level model which uses a shallow intra-word transformer encoder to learn word representations from characters and a deep inter-word encoder that attends to the entire word sequence. As a result, (1) it gives direct access to characters without requiring long sequence lengths, (2) it preserves explicit word boundaries, and (3) it allows for an open vocabulary.

For the intra-word encoder, we use a sequence length of 16 which is long enough to cover the vast majority of words in our training data. While \citet{sun-etal-2023-characters} truncate words that exceed the maximum sequence length of the intra-word encoder, we instead split them into multiple subwords to avoid any loss of information. For the inter-word encoder we use a maximum sequence length of 512. Because the intra-word encoder is limited to characters within the same word and the inter-word encoder operates on word sequences, this approach is computationally more efficient than a vanilla character model, and even approaches the performance of subword-based models \citep{sun-etal-2023-characters}.

The input to the intra-word encoder is produced by encoding each word into a sequence of character tokens, with a special \texttt{[WORD\_CLS]} token inserted at the beginning of each word. The contextualized \texttt{[WORD\_CLS]} embeddings from the intra-word encoder are then used as the word representations for the inter-word encoder.

We create a character tokenizer for each language using a character vocabulary consisting of all the unique characters found in the training data for that language.  Any unseen characters encountered in the validation or test data are replaced with a special \texttt{[UNK]} token. Table \ref{tab:vocab} shows the vocabulary sizes for each language, including special tokens. The character vocabularies are typically quite small, with the notable exception of Classical Chinese (lzh), where most of the tokens in the training data are single characters. We experimented with several decomposition methods, inspired by the work of \citet{si-etal-2023-sub} on sub-character tokenization for Chinese. However, we were unable to improve performance on our downstream tasks, so we opted to use the same character tokenization method for all languages.

\subsection{Hierarchical Encoder-only Models}

\begin{figure*}[ht]
	\centering
	\includegraphics[width=16cm]{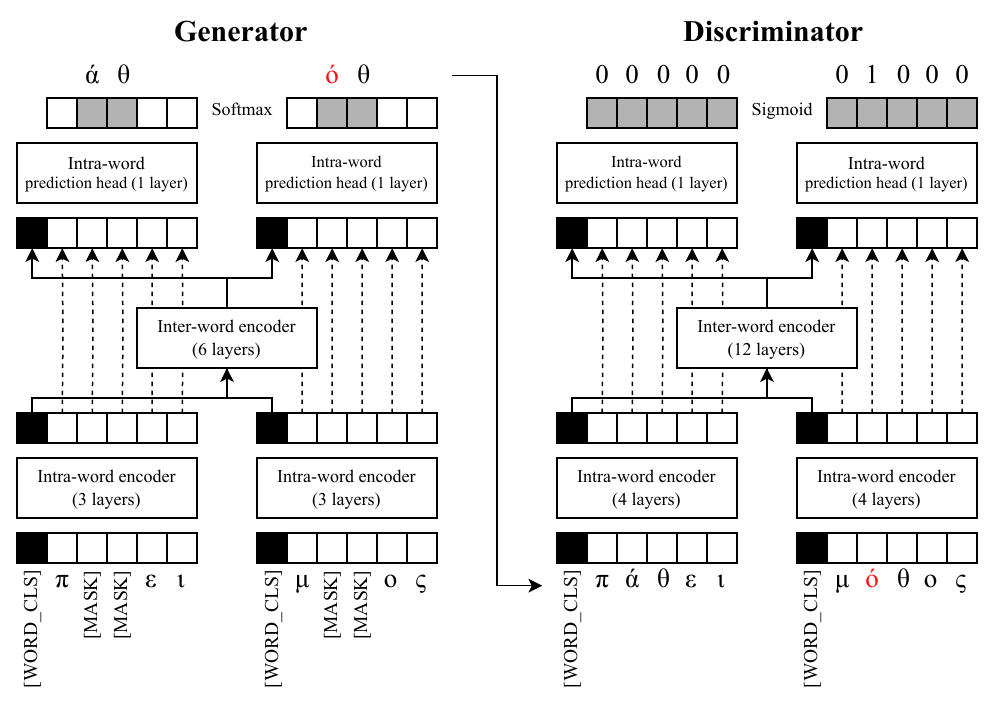}
	\caption{HLM-DeBERTa architecture with RTD pre-training. Input text is \enquote{\textgreek{πάθει μάθος}}.}
	\label{fig:hlm_rtd}
\end{figure*}

To conduct PoS and morphological tagging, we rely on an encoder that generates the necessary word embeddings for classification. Our encoder models build on a modified implementation of DeBERTa-V3 \citep{he2023debertav}, combining the advantages of HLM with the DeBERTa architecture. The intra- and inter-word modules are implemented as two separate DeBERTa encoders, utilizing disentangled attention \citep{he2021deberta} and relative position encoding.

\paragraph{Replaced Token Detection.} For the pre-training task we use replaced token detection (RTD), originally proposed by \citet{clark2020electra}. RTD uses a generator model to generate corrupted input sequences and a discriminator to distinguish between the original and corrupted tokens. After training, the generator is discarded and the discriminator is fine-tuned for downstream tasks. In our experiments, when applying RTD pre-training, we achieve slightly better performance on our downstream tasks compared to masked language modeling (MLM) as the pre-training task. Following previous work \citep{he2023debertav, clark2020electra}, we use a generator
with roughly half the model parameters compared to the discriminator. We train a monolingual model for each language for 30 epochs. Further pre-training does not improve performance on downstream tasks.

We utilize DeBERTa-V3's gradient-disentangled embedding sharing (GDES),
which allows the embedding gradients from the generator to flow directly
to the discriminator, but not vice versa. This results in more stable training compared to the vanilla embedding sharing (ES) used by ELECTRA \citep{clark2020electra}, which allows the gradients to flow in both directions.

\paragraph{Masking Strategy.} We use character-level masking to allow for open-vocabulary language modeling. The character token sequence is restored by concatenating the character representations from the intra-word module with the word representations from the inter-word module, replacing the initial \texttt{[WORD\_CLS]} with the contextualized representation. We follow the original HLM approach for the language modeling prediction head: an additional single-layer intra-word transformer module followed by a simple feed-forward network. A softmax layer is used for the generator's output distribution and a sigmoid layer is used for the discriminator. The relative position embedding matrix is shared between the initial intra-word encoder and the intra-word language modeling head. Figure \ref{fig:hlm_rtd} shows an overview of our architecture for RTD pre-training.

We compare the following masking strategies:

\begin{itemize}
    \item Whole-word masking: mask the characters in 15\% of the words (original HLM approach),
    \item Character masking: randomly mask 15\% of the characters,
    \item Character n-gram masking: mask random spans of 1-4 characters until 15\% of the characters are masked.
\end{itemize}

Through experimentation we found that character n-gram masking performed best for our
downstream tasks, by a small margin. Random character masking performed similarly to
whole-word-masking. We hypothesize that it is too difficult for the model
to learn to predict whole words from the small training corpora. Conversely, random
character masking is too easy, as MLM pre-training accuracy reaches high levels very quickly.

\subsection{Character-level Encoder-decoder Models}
\label{subsec:encdec}
While encoder-only models are very effective for classification tasks, lemmatization is most naturally treated as a sequence-to-sequence problem, where the inflected form is \enquote{translated} to its lemma. We therefore choose to train an encoder-decoder model that handles sequence-to-sequence tasks naturally. Specifically, we train a T5 model for each language \citep{2020t5} using the nanoT5 library \citep{nawrot-2023-nanot5} and the t5-v1\_1-base configuration. In lemmatization, our aim is to prioritize the characters within a word, rather than focusing on a detailed understanding of contextualized words (see \cref{subsec:lemmatization} for our approach). Moreover, extending a hierarchical structure to (encoder-)decoder models like T5 is not straightforward. Therefore, we employ character tokenization in the T5 models for lemmatization.

\section{Using our PLMs for Downstream Tasks}

Many systems focusing on Universal Dependencies, often introduced in shared tasks, utilize cross-lingual transfer and multi-task learning. For instance, UDPipe \citep{udpipe}, which employs multilingual BERT, is fine-tuned on specific treebanks for PoS tagging, morphological tagging, lemmatization, and dependency parsing. UDify \citep{kondratyuk-straka-2019-75} learns these tasks for 75 languages in one model. 

Given that in our setting cross-lingual transfer is excluded, we investigate multi-task learning as a remaining option to leverage additional training signals for resource-poor languages.

\subsection{Morphological Tagging}
Following \citet{riemenschneider-frank-2023-exploring}, we treat morphological tagging as a multi-task-classification problem, where every token is processed through \(k\) classification heads, corresponding to each possible morphological feature in a dataset. Whenever a feature is missing in a token, the model is trained to predict a class indicating the feature's absence. 

To represent a token, the HLM architecture yields two kinds of embeddings: those derived from the intra-word encoder, informed by a word's characters but not by other sentence words, and those that are contextualized by surrounding tokens. In line with \citet{sun-etal-2023-characters} as well as earlier work \citep{clark-etal-2022-canine,plank-etal-2016-multilingual}, we concatenate these embeddings to create a unified final word representation.

We use a simple feed-forward network followed by a softmax function on top of the last hidden state of this word representation.
The final loss is computed as:
\[
\mathcal{L}_{\text{morph}} = \frac{1}{k} \sum_{m=0}^{k-1}  \mathcal{L}_m\,
\]
where \(k\) is the number of morphological features.

We further extended the multi-task framework to include additional related tasks, hypothesizing that obtaining training signals from auxiliary tasks could improve the model's capabilities, particularly under our low-resource conditions. To this end, we incorporated tasks such as dependency parsing and PoS tagging. Contrary to our expectations, this approach led to slower convergence and did not provide any performance benefits, occasionally even producing marginally inferior results. We discuss these findings in \cref{sec:negresults}.

\subsection{PoS Tagging}
Analogous to our approach in morphological tagging, we represent each token by concatenating its intra- and inter-word embeddings, followed by a classification head. However, in contrast to morphological tagging, we notice slight improvements when the model is also tasked with predicting morphological features. Thus, we determine the loss as \(\mathcal{L}_{\text{UPoS}} + \mathcal{L}_{\text{morph}}\), disregarding the morphological tagging predictions during inference. 

\subsection{Lemmatization}
\label{subsec:lemmatization}

As outlined in \cref{subsec:encdec}, lemmatization is most naturally treated as a sequence-to-sequence problem, where the form to be lemmatized is transduced into its lemma, which is why we propose using a T5 model for this task. Ideally, our model should receive the word to be lemmatized in its original context, while marking the word to be lemmatized, similar to the approach used by \citet{riemenschneider-frank-2023-exploring}. For instance, given the input sequence \textgreek{ξύνοιδα} \texttt{[SEP]} \textgreek{ἐμαυτῷ} \texttt{[SEP]} \textgreek{οὐδὲν ἐπισταμένῳ}, the model would be expected to predict the lemma of \textgreek{ἐμαυτῷ}, which is \textgreek{ἐμαυτοῦ}. This approach would enable us to train the model in an end-to-end fashion, allowing it to autonomously learn the relevant information directly from the word within its contextual surroundings.

However, this training method is prohibitively expensive, requiring repeated passes through the model, once for each token in the sentence. Moreover, we noted that the models exhibited exceptionally slow convergence. Allowing the model to predict lemmata for all words in a sentence in a single forward pass mitigates the computational challenges, as it requires only one pass per sentence per epoch. Yet, this strategy still encounters problems with very slow, and at times nonexistent, convergence, while also introducing new challenges for the model, particularly in assigning exactly one lemma to each token accurately.

Therefore, we adopt a pipeline approach, following \citet{wrobel-nowak-2022-transformer}, by providing the model with the inflected form and its corresponding UPoS tag. For training purposes, we use the gold UPoS tag, whereas for inference we rely on the UPoS tag as predicted by our HLM-DeBERTa model. We predict lemmata using beam search with a beam width of 20, restricting the maximum sequence length to 30.

\begingroup
\renewcommand{\arraystretch}{1.25}
\begin{table*}[!htbp]
    \centering
    \resizebox{\linewidth}{!}{
    \begin{threeparttable}
    \begin{tabular}{llrrrrrrrrrrrrr}
        \toprule
        &Language: & chu & cop & fro & got & grc & hbo & isl & lat & latm & lzh & ohu & orv & san\\
        \cmidrule{1-1} \cmidrule(l{5pt}){2-15}
        \multicolumn{14}{l}{\textbf{Morphological Tagging}} \\
        \midrule
        \multirow{2}{*}{Constrained} & Ours & \textbf{96.04} & \textbf{98.60} & \textbf{97.87} & \textbf{95.32} & \textbf{97.46} & \underline{\textbf{97.46}} & \textbf{95.29} & \textbf{95.17} & \textbf{98.68} & \textbf{95.52} & \textbf{96.30} & \textbf{95.00} & \textbf{91.58} \\
        & Team 21a           & 94.06 & 80.47 & 94.08 & 93.96 & 96.50 & 71.20 & 94.79 & 93.31 & 97.98 & 85.98 & 94.64 & 92.16 & 90.00\\
        & Baseline & 
        85.07 & 47.41 & 28.27 & 18.95 & 25.10 & 42.78 & 35.83 & 18.17 & 30.94 & 43.58 & 23.20 & 25.55 & 08.34\\
        \hdashline
        \multirow{2}{*}{Unconstrained} & UDParse & \underline{\textbf{96.49}} & \underline{\textbf{98.88}} & \underline{\textbf{98.33}} & \underline{\textbf{96.23}} & \underline{\textbf{97.78}} & \textbf{97.05} & \underline{\textbf{95.92}} & \underline{\textbf{96.66}} & \underline{\textbf{98.83}} & \underline{\textbf{96.24}} & \underline{\textbf{96.62}} & \underline{\textbf{95.16}} & \underline{\textbf{92.60}} \\
        & TartuNLP  & 67.14 & 74.86 & 98.01 & 92.40 & 97.33 & 95.14 & 95.53 & 95.91 & \underline{\textbf{98.83}} & 88.75 & 75.62 & 80.00 & 86.33 \\
        \midrule
        \multicolumn{14}{l}{\textbf{PoS Tagging}} \\
        \midrule
        \multirow{2}{*}{Constrained} &Ours & \textbf{96.57} & \textbf{96.92} & \textbf{93.10} & \textbf{95.41} & \textbf{96.39} & \textbf{96.68} & \textbf{96.08} & \textbf{95.54} & \textbf{98.43} & \textbf{92.92} & \textbf{95.98} & \textbf{94.46} & \textbf{89.71} \\
        &Team 21a           & 94.62 & 42.65 & 85.14 & 93.48 & 93.49 & 27.26 & 93.85 & 92.43 & 94.41 & 81.79 & 94.42 & 91.23 & 87.32 \\
        & Baseline & 93.36 & 94.98 & 91.57 & 93.73 & 90.33 & 94.07 & 94.00 & 92.39 & 97.22 & 90.91 & 93.59 & 90.33 & 89.37\\
        \hdashline
        \multirow{2}{*}{Unconstrained} & UDParse & \underline{\textbf{97.00}} & \underline{\textbf{97.33}} & \underline{\textbf{96.01}} & \underline{\textbf{96.47}} & \underline{\textbf{96.49}} & \underline{\textbf{97.84}} & \underline{\textbf{96.88}} & \underline{\textbf{96.83}} & \underline{\textbf{98.79}} & \underline{\textbf{93.76}} & \underline{\textbf{96.71}} & \underline{\textbf{94.99}} & \underline{\textbf{90.02}} \\
        & TartuNLP  & 66.35 & 60.99 & 94.51 & 92.72 & 95.72 & 94.15 & 96.67 & 95.86 & \underline{\textbf{98.79}} & 83.28 & 75.14 & 75.67 & 83.83\\
        \midrule
        \multicolumn{14}{l}{\textbf{Lemmatization}} \\
        \midrule
        \multirow{2}{*}{Constrained} & Ours & \underline{\textbf{94.49}} & 95.07 & \textbf{92.63} & \textbf{93.31} &  \underline{\textbf{94.08}}& \textbf{97.29} & \textbf{96.63} & \textbf{96.00} & \textbf{98.46} & 99.18 & 85.92 & \underline{\textbf{90.09}} & \textbf{84.59}\\
        & Team 21a           & 79.59 & 46.32 & 83.32 & 90.79 & 88.30 & 61.75 & 94.58 & 92.35 & 97.22 & \textbf{99.84} & 69.97 & 78.44 & 83.21\\
        & Baseline & 89.60 & \textbf{95.74} & 91.93 & 91.95 & 91.06 & 95.28 & 93.78 & 92.08 & 97.03 & 98.81 & \underline{\textbf{89.43}} & 84.44 & 84.24 \\
        \hdashline
        \multirow{2}{*}{Unconstrained} & UDParse & 59.56 & 74.78 & 92.47 & 92.81 & \textbf{94.02} & 96.85 & \underline{\textbf{97.96}} & \textbf{96.74} & \underline{\textbf{98.91}} & \underline{\textbf{99.96}} & 63.43 & 68.55 & 88.10 \\
        & TartuNLP & \textbf{92.70} & \underline{\textbf{98.28}} & \underline{\textbf{95.11}} & \underline{\textbf{95.41}} & 93.39 & \underline{\textbf{98.15}} & 97.23 & \underline{\textbf{96.99}} & 98.69 & 99.91 & \textbf{86.91} & \textbf{89.23} & \underline{\textbf{91.48}}  \\
        \bottomrule
    \end{tabular}
 	\end{threeparttable}}

    \caption{Results on \textit{SIGTYP 2024 Shared Task on Word Embedding Evaluation for Ancient and Historical Languages}. We mark the winner of each subtask in \textbf{bold} and \underline{underline} the overall winner. See \cref{sec:data} for language identifiers.}
    \label{tab:all_results}
\end{table*}
\endgroup

\section{Results}
Our results are computed using the SIGTYP 2024 official evaluation script.\footnote{\url{https://github.com/sigtyp/ST2024/blob/main/scoring_program_constrained.zip}.} 
The script computes PoS tagging scores as the unweighted average of the accuracy and the F\(_\text{1}\) score. For morphological tagging, it computes the averaged accuracy across each token, with deductions for any feature categories predicted by the model but absent in the label. The lemmatization scores are the unweighted average of the accuracy@1 and the accuracy@3.

We report our results in \cref{tab:all_results} and provide dataset statistics in \cref{sec:data}.  In \textbf{PoS} and \textbf{morphological tagging}, our system emerges as the winner of the constrained task. Its performance is consistently almost on-par with that of the unconstrained task winner, being only 0.69 percentage points lower on average. A notable outlier is seen in Old French (fro) PoS tagging, where our system falls short by 3 percentage points. This performance difference might be linked to the small size of the Old French corpus in the treebank,  although our model generally shows strong performance in learning from small datasets, as demonstrated by its robust performance in other datasets of similar size, such as Ancient Hebrew (hbo), Gothic (got), and Vedic Sanskrit (san).

Results in \textbf{lemmatization} display greater diversity, likely due to the differing architectures in participants' approaches.  Our model achieves 99.18\% in Classical Chinese (lzh), a language where distinct lemmata do not really exist, usually turning the task into mere form replication. This score, though precise, is somewhat lower than the near-perfect range of 99.81 to 99.96\% achieved by the other methods in the shared task.

\section{Negative Results}
\label{sec:negresults}

\paragraph{Multi-task Learning.} We hypothesized that 
a model simultaneously doing PoS tagging, morphological tagging and dependency parsing could benefit from the training signals of related tasks.\footnote{For dependency parsing, we adopt the head selection method as described by \citet{zhang-etal-2017-dependency}.} However, this approach did not significantly improve morphological ana\-lysis and resulted in longer training times due to slower convergence.  On the other hand, jointly performing morphological and PoS tagging in a multi-task learning setup 
yielded minor improvements in PoS tagging. We believe that including PoS information offers little extra insight to the model for morphological tagging and simultaneously pressures it to form representations apt for PoS tagging. Conversely, enriching the coarser PoS tagging task with morphological labels provides the model with useful additional insights. Furthermore, our dependency parsing technique differs from the more direct classification approach used in PoS and morphological tagging, potentially leading to instabilities during training.

\paragraph{Tall Models.}
\citet{10.5555/3618408.3620029} found that transformers with a narrower and deeper architecture might surpass the performance of similarly sized models in masked language modeling tasks. Inspired by this finding, we experimented with doubling the number of layers to 24 while reducing the hidden size from 768 to 512 and the number of attention heads from 12 to 8. However, although this adjustment seemed to yield a marginal improvement in pre-training with MLM, it did not result in any performance changes when training with RTD.

\section{Conclusion}
We present our approach for the SIGTYP 2024 shared task on historical language analysis. Our method employs a hierarchical transformer that first focuses on a word's characters, applying self-attention to generate initial word embeddings. These embeddings are then further developed by integrating the contextual information from surrounding words. We pre-train HLM-DeBERTa-V3 and T5 models with small datasets of historical texts. The character-based methodology of our architecture yielded promising results, effectively leveraging the available data. Contrary to our expectations, the implementation of multi-task learning had only a negligible effect on enhancing our models' performance.

\section*{Acknowledgements}

We thank Anette Frank for her helpful suggestions and her constructive feedback on our paper. We are deeply grateful to Fabian Strobel for his support and the valuable pointers he provided.

\clearpage
\bibliography{anthology,custom}
\clearpage
\appendix

\section{Pre-Training Details}
\label{sec:training}

\begin{table}[htbp]
	\centering
	\resizebox{\linewidth}{!}{
	\begin{tabular}{lrr}
        \toprule
		\textbf{Parameter} & \textbf{Generator} & \textbf{Discriminator} \\
        \midrule
        Activation & GELU & GELU \\
        Hidden Dropout & 0.1 & 0.1 \\
        Initializer Range & 0.02 & 0.02 \\

        \midrule
        \textbf{Intra-word encoder} \\
        \midrule
        Layers & 3 & 4 \\
        Hidden Size & 768 & 768 \\
        Intermediate Size & 1536 & 1536 \\
        Attention Heads & 12 & 12 \\

        \midrule
        \textbf{Inter-word encoder} \\
        \midrule
        Layers & 6 & 12 \\
        Hidden Size & 768 & 768 \\
        Intermediate Size & 3072 & 3072 \\
        Attention Heads & 12 & 12 \\
        \bottomrule
	\end{tabular}
    }
	\caption{HLM-DeBERTa hyperparameters.}
\end{table}

\begin{table}[htbp]
	\centering
	%\small
	\begin{tabular}{lr}
        \toprule
		\textbf{Parameter} & \textbf{Value} \\
		\midrule
        Optimizer & Adam \\
        Weight Decay & 0.01 \\
        Batch Size & 16 \\
        Learning Rate & 1e-5 \\
        Learning Rate Scheduler & constant \\
        Epochs & 30 \\
        Warmup Proportion & 0.1 \\
        Mask Percentage & 15\% \\
        Max Sequence Length (words) & 512 \\
        Max Word Length (chars) & 16 \\
        \bottomrule
	\end{tabular}
	\caption{HLM-DeBERTa pre-training hyperparameters.}
\end{table}

\begin{table}[htbp]
	\centering
	\begin{tabular}{lrr}
        \toprule
		\textbf{Parameter} & \textbf{Encoder} & \textbf{Decoder} \\
        \midrule
        Activation & GEGLU & GEGLU \\
        Hidden Dropout & 0.0 & 0.0 \\
        Layers & 12 & 12 \\
        Hidden Size & 768 & 768 \\
        Intermediate Size & 2048 & 2048 \\
        Attention Heads & 12 & 12 \\
        \bottomrule
	\end{tabular}
	\caption{T5 hyperparameters.}
\end{table}

\begin{threeparttable}[htbp]
	\centering
	%\small
	\begin{tabular}{lr}
        \toprule
		\textbf{Parameter} & \textbf{Value} \\
		\midrule
        Optimizer & AdamWScale\tnote{*} \\
        Weight Decay & 0.0 \\
        Batch Size & 16 \\
        Learning Rate & 1e-5 \\
        Learning Rate Scheduler & cosine \\
        Epochs & 100 \\
        Warmup Steps & 1000 \\
        Mask Percentage & 15\% \\
        Max Sequence Length & 512 \\
        Mean Noise Span Length & 3 \\
        \bottomrule
	\end{tabular}
	\caption{T5 pre-training hyperparameters.}
 
 \begin{tablenotes}
 		\item[*] We use the customized AdamW implementation of nanoT5 \citep{nawrot-2023-nanot5} that is augmented by RMS scaling.
 \end{tablenotes}
\end{threeparttable}
\FloatBarrier

\section{Fine-tuning Details}
\label{sec:fine}
\begin{table}[htbp]
	\centering
	%\small
	\begin{tabular}{lr}
        \toprule
		\textbf{Parameter} & \textbf{Value} \\
		\midrule
        Optimizer & AdamW \\
        Weight Decay & 0.01 \\
        Batch Size & 16 \\
        Learning Rate & 2e-5 \\
        Learning Rate Scheduler & linear \\
        Early Stopping Patience & 10 \\
        \bottomrule
	\end{tabular}
	\caption{HLM-DeBERTa fine-tuning hyperparameters.}
\end{table}

\begin{table}[htbp]
	\centering
	%\small
	\begin{tabular}{lr}
        \toprule
		\textbf{Parameter} & \textbf{Value} \\
		\midrule
        Optimizer & AdamW \\
        Weight Decay & 0.01 \\
        Batch Size & 16 \\
        Learning Rate & 1e-3 \\
        Learning Rate Scheduler & linear \\
        Early Stopping Patience & 10 \\
        \bottomrule
	\end{tabular}
	\caption{T5 fine-tuning hyperparameters.}
\end{table}

\onecolumn

\FloatBarrier
\section{Dataset Statistics}
\label{sec:data}

\begin{table}[!htbp]
    \centering
    \resizebox{\linewidth}{!}{
    \begin{tabular}{llllrrrrrr}
    \toprule
      \textbf{Language} & \textbf{Code} & \textbf{Family} & \textbf{Script} & \textbf{Train Tok.} & \textbf{Valid Tok.} & \textbf{Test Tok.} & \textbf{Train Sent.} & \textbf{Valid Sent.}& \textbf{Test Sent.}\\
      \midrule
       Ancient Greek  & grc & Indo-European & Greek & \num{334043}  &\num{41905} & 	\num{41046} & \num{24800} & \num{3100} & \num{3101}\\
       Ancient Hebrew  & hbo & Afro-Asiatic & Hebrew & \num{40244}  &\num{4862} & 	\num{4801} & \num{1263} & \num{158} & \num{158}\\
       Classical Chinese  & lzh & Sino-Tibetan & Hanzi & \num{346778}  &\num{43067} & 	\num{43323} & \num{68991} & \num{8624} & \num{8624}\\
       Coptic  & cop & Afro-Asiatic & Egyptian & \num{57493}  &\num{7272} & 	\num{7558} & \num{1730} & \num{216} & \num{217}\\
       Gothic & got & Indo-European & Latin & \num{44044}  &\num{5724} & 	\num{5568} & \num{4320} & \num{540} & \num{541}\\
       Medieval Icelandic  & isl & Indo-European & Latin & \num{473478}  &\num{59002} & 	\num{58242} & \num{21820} & \num{2728} & \num{2728}\\
       Classical \& Late Latin & lat & Indo-European & Latin & \num{188149}  &\num{23279} & 	\num{23344} & \num{16769} & \num{2096} & \num{2097}\\
       Medieval Latin  & latm & Indo-European & Latin & \num{599255}  &\num{75079} & 	\num{74351} & \num{30176} & \num{3772} & \num{3773}\\
       Old Church Slavonic  & chu & Indo-European & Cyrillic & \num{159368}  &\num{19779} & 	\num{19696} & \num{18102} & \num{2263} & \num{2263}\\
       Old East Slavic  & orv & Indo-European & Cyrillic & \num{250833}  &\num{31078} & 	\num{32318} & \num{24788} & \num{3098} & \num{3099}\\
       Old French & fro & Indo-European & Latin & \num{38460}  &\num{4764} & 	\num{4870} & \num{3113} & \num{389} & \num{390}\\
       Vedic Sanskrit & san & Indo-European & Latin (transcr.) & \num{21786}  &\num{2729} & 	\num{2602} & \num{3197} & \num{400} & \num{400}\\
       Old Hungarian  & ohu & Finno-Ugric & Latin & \num{129454}  &\num{16138} & 	\num{16116} & \num{21346} & \num{2668} & \num{2669}\\
    \bottomrule
    \end{tabular}}
    \caption{Dataset statistics.}
    \label{tab:dataset_statistics}
\end{table}
\end{document}